\documentclass{article}

\usepackage{spconf,amsmath,amssymb,graphicx}

\usepackage{booktabs}
\usepackage{multirow}
\usepackage{siunitx}
\usepackage[hidelinks]{hyperref}
\usepackage{cite}
\usepackage{listings}
\usepackage{microtype}
\usepackage[utf8]{inputenc}
\DeclareUnicodeCharacter{0080}{}

\usepackage{array}      
\usepackage{tabularx}
\usepackage{siunitx}
\title{MMSE-Calibrated Few-Shot Prompting for Alzheimer's Detection}

\name{Jana Sweidan$^{1}$ \quad Mounim A.~El-Yacoubi$^{1}$ \quad Nasredine Semmar$^{2}$}

\address{%
$^{1}$ T\'el\'ecom SudParis (Samovar), Institut Polytechnique de Paris, Palaiseau, France\\
$^{2}$ CEA (List), Universit\'e Paris-Saclay, Palaiseau, France\\
\texttt{\{jana\_sweidan,\,mounim.el\_yacoubi\}@telecom-sudparis.eu}\\
\texttt{nasredine.semmar@cea.fr}
}

\begin{document}
\ninept 
\maketitle

\begin{abstract}

Prompting large language models is a training-free method for detecting Alzheimer’s disease from speech transcripts. Using the ADReSS dataset, we revisit zero-shot prompting and study few-shot prompting with a class-balanced protocol using nested interleave and a strict schema, sweeping up to 20 examples per class. We evaluate two variants achieving state-of-the-art prompting results. (i) MMSE-Proxy Prompting: each few-shot example carries a probability anchored to Mini-Mental State Examination bands via a deterministic mapping, enabling AUC computing; this reaches 0.82 accuracy and 0.86 AUC (ii) Reasoning-augmented Prompting: few-shot examples pool is generated with a multimodal LLM (GPT-5) that takes as input the Cookie Theft image, transcript, and MMSE to output a reasoning and MMSE-aligned probability; evaluation remains transcript-only and reaches 0.82 accuracy and 0.83 AUC. To our knowledge, this is the first ADReSS study to anchor elicited probabilities to MMSE and to use multimodal construction to improve interpretability.

\end{abstract}


\section{Introduction}\label{sec:intro}

Alzheimer’s disease (AD) is the most common cause of dementia and a growing public-health challenge worldwide~\cite{adi_facts}. Diagnostic practice increasingly uses biomarker-based frameworks and staged criteria with an emphasis on earlier detection and longitudinal tracking~\cite{mckhann2011_niaaa,jack2018_niaaa,jack2024_revised}. Because earlier identification can inform counseling and care planning at prodromal stages~\cite{petersen1999_mci}, there is demand for scalable, low-burden markers that fit routine clinical workflows. Language meets these constraints: it is non-invasive, inexpensive to collect, and sensitive to cognitive change. Alterations in connected speech in AD span lexical diversity, syntax, discourse coherence and informativeness, disfluencies, and timing~\cite{fraser2016jad,fraser2019automated,ahmed2013_connected}. The Cookie Theft picture from the Boston Diagnostic Aphasia Examination standardizes elicitation and foregrounds discourse-level impairments~\cite{goodglass2001_bdae}. To enable comparability, ADReSS 2020 released a balanced, age- and sex-matched subset of Pitt Cookie Theft recordings with fixed splits and standardized preprocessing that many later pipelines adopted~\cite{ADReSStask2020,martinc2021fan,guo2021chasm}. Beyond English, Cookie Theft descriptions have been collected in several languages indicating portability of the paradigm~\cite{kokkinakis2018swedish,pereztoro2022spanish,li2019mandarin}. In clinical practice, the Mini-Mental State Examination (MMSE) provides a 0–30 severity scale~\cite{folstein1975mmse}. Earlier work emphasized acoustic and multimodal cues. With recent progress in large pre-trained language models (LLMs), we focus on transcript-only detection to exploit language modeling strengths. 
Prompting keeps model weights fixed and steers behavior with instructions and a few labeled demonstrations (few-shot examples) at inference time, avoiding weight updates but remaining sensitive to exemplar choice and ordering, probability elicitation, and output formatting. A classifier-head-free alternative replaces direct prompting with likelihood-based decision rules such as paired perplexity, but recent implementations typically fine-tune separate class-conditional language models and still depend on preprocessing and threshold selection~\cite{xiao2025ppl},  which reduces the simplicity advantage of prompting-only.  Training-based adaptation, including prompt fine-tuning and parameter-efficient variants, can be accurate on ADReSS but require optimization, hyperparameter selection, and careful regularization in a data-scarce regime, which raises compute and deployment burden and risks overfitting~\cite{liu2024promptft,zheng2025prompt,park2025cot}. Comparative analyses on narrative speech suggest that fine-tuned encoders and prompted LLMs capture different strengths, reinforcing the value of carefully designed prompting protocols rather than defaulting to training~\cite{mekulu2025medrxiv}.
 Within prompting on Cookie Theft transcripts, two baselines are most relevant. Botelho et al. evaluate zero- and few-shot prompting on ADReSS transcripts and report that Mistral-7B-Instruct is the strongest prompting-only baseline with accuracy around 75\% on the test set; however they record six generation failures, defined as outputs that hallucinate or do not produce a parsable AD or healthy control(HC) decision under the required schema, and they count these fails as HC, a policy that can bias error profiles and complicate reproducibility~\cite{botelho2024macro}. A subsequent study examines zero- and few-shot prompting with LLaMA-3.1 and a vision-language variant and finds that adding vision brings no clear benefit on text transcripts, consistent with the task’s primarily linguistic signal~\cite{cl4health2025}. Related work that is not prompting-only uses LLMs to derive features that are then fed to supervised classifiers~\cite{liu2024promptft,mo2024fewshot, botelho2024macro}; these approaches can work well but reintroduce optimization steps, thresholds, and model-selection choices that prompting-only explicitly avoids. Prompting studies establish it as a viable, training-free approach on ADReSS manual transcripts, but they leave some gaps such as systematic exploration of context size beyond two demonstrations per class, and anchoring of elicited probabilities to a clinical scale.
We build directly on the strongest prompting baseline on ADReSS manual transcripts (Mistral)~\cite{botelho2024macro} and revisit zero- and few-shot prompting targeting both performance and reproducibility. We use a nested, class-alternating interleave that keeps the exemplar order class-balanced at every step, mitigating order and recency effects. We require a strict forced-decision JSON schema to eliminate parsing-related scoring artifacts, directly addressing failure conventions reported previously. We elicit probabilities on an MMSE-anchored scale so that ROC–AUC can be reported without post-hoc calibration and with monotonicity to clinical severity. Unlike prior Cookie Theft prompting that typically uses up to two demonstrations per class, we sweep from $k$=0 to $k$=20 exemplars per class and report accuracy and AUC as a function of context size. In addition, we construct a reasoning-augmented exemplar pool once with a multimodal LLM to attach concise, transcript-grounded rationales that enhance interpretability of the LLM decisions.
Recently a new paper has been published on a retrieval-based alternative, Delta-KNN, which builds a held-out Delta Matrix of one-shot gains combined with nearest neighbors from external embeddings; it improves few-shot prompting but introduces a selection-time pipeline that depends on external embedding model and tuned hyperparameters~\cite{li2025delta}. Our protocol avoids external retrievers and delta matrices and mainly uses a global exemplar pool for all transcripts during inference.

Accordingly, our contributions are as follows: (1) MMSE-Proxy Prompting, a deterministic class-conditional mapping from MMSE to AD probability that enables threshold-free ROC–AUC; (2) Reasoning-augmented Prompting, a one-time construction of exemplar rationales with a multimodal LLM to improve interpretability; (3) the first systematic sweep from $k{=}0$ to $k{=}20$ examples per class on Cookie Theft transcripts under a controlled setup; and (4) a balanced few-shot prompting protocol with nested interleave and a strict forced-decision schema that yields reliable outputs.

\section{Dataset and Preprocessing}\label{sec:data}
We use the ADReSS~2020 benchmark, which is a balanced subset of DementiaBank’s Pitt Cookie Theft picture descriptions and was specifically released for standardized AD detection research \cite{ADReSStask2020}. The corpus originates from DementiaBank within TalkBank and includes audio plus manual CHAT transcripts of 156 speakers, 78 AD and 78 HC \cite{ADReSStask2020,dementiabank}. We follow the official split exactly: 108 training recordings and 48 test recordings, matched for age and sex at the cohort level (Train per class: 24 male, 30 female; Test per class: 11 male, 13 female). We keep the official train/test split to avoid any leakage \cite{ADReSStask2020}. Language from Cookie Theft picture descriptions is a well-established clinical signal in AD, and transcript features can reliably distinguish AD from controls \cite{fraser2016jad,botelho2024macro}.

We prioritize manual CHAT transcripts to retain diagnostic cues like disfluencies, repetitions, retracing, reduced forms, pauses, and unintelligible spans that are characteristic of AD discourse \cite{fraser2016jad,botelho2024macro}. Off-the-shelf ASR systems often struggle with disfluent or disordered speech and can omit or normalize exactly these cues, which remains an active challenge despite recent progress \cite{teleki2024disfluency,mujtaba2024inclusive,gohider2024disordered}. Elevated ASR error rates are also documented for older adults and pathological speech, which are central to AD cohorts \cite{vipperla2010ageing,gohider2024disordered}. 


\noindent\textbf{Transcript extraction:}
We parse CHAT files and keep only participant speech marked by the \texttt{*PAR:} tier, excluding investigator turns and dependent metadata tiers as specified in the CHAT manual \cite{chatmanual}. We remove structural control codes while preserving clinical content; for example, we convert retracing \texttt{<what are> [/]} to the readable repetition “what are what are,” and we render disfluency markers like \texttt{\&uh} and \texttt{\&um} as “uh” and “um,” following CHAT conventions. We preserve unintelligible spans as \texttt{xxx} and keep simple phonological intrusions such as \texttt{\&k} or \texttt{\&sh} as “k” and “sh”. We also make pauses explicit by mapping CHAT pause marks \texttt{(.), (..), (...)} to “(short pause),” “(medium pause),” and “(long pause)” so that non-lexical timing information remains visible to text models.

\section{Methods}\label{methods}
In this section, we begin by introducing our prompting strategy and example selection (subsection \ref{sec:protocol0}). Then, we explain the two prompting variants on ADReSS transcripts, starting with the  MMSE-Proxy Prompting (subsection \ref{sec:protocol1}) and moving to Reasoning-augmented examples constructed with GPT-5 (subsection \ref{sec:protocol2}). Finally, we describe the models used and the evaluation metrics (subsection \ref{sec:models_eval}).

\subsection{Prompting and example selection}\label{sec:protocol0}
We adopt the best-performing Cookie Theft prompt of Botelho et\,al.~\cite{botelho2024macro} on manual transcripts in a compact ChatML-style format. The instruction frames the model as a clinical evaluator of Alzheimer’s disease (AD) from Cookie Theft transcripts and requires a single JSON object with 3 fields: {comment} (a concise string for reasoning), alzheimers\_prediction (YES or NO), and probability\_score (a float in $[0,1]$ giving the likelihood of AD). This score replaces the high/low confidence field in \cite{botelho2024macro}.  The instruction includes a forced-decision clause (“choose the more likely class; never output MAYBE, UNCERTAIN...”\,).  Outputs are counted as failures if the JSON cannot be parsed or probability\_score is missing or out of range. 
We prepend $k$ examples per class (AD, HC) to the instruction. 

Examples are selected by a seed-deterministic within-class shuffle and then ordered with a class-alternating interleave \((a_1,\allowbreak h_1,\allowbreak \ldots,\allowbreak a_k,\allowbreak h_k)\), where \(a_i\) and \(h_i\) denote AD and HC exemplars. Decoder-only LLMs are order- and recency-sensitive~\cite{liu2023_lost_middle}, so naive random concatenation can change the early context as \(k\) grows. Because the interleave is nested, each increase in \(k\) appends new exemplars at the end and leaves the existing context unchanged, which reduces positional bias and seed variance and isolates the effect of \(k\).

\subsection{MMSE-Proxy Prompting}\label{sec:protocol1}
 For MMSE-Proxy Prompting, each example carries a probability produced by a fixed, predefined monotone mapping from MMSE to AD probability; the mapping is deterministic, contains no learned parameters, and is aligned to published MMSE banding that maps MMSE to Clinical Dementia Rating (CDR) stages~\cite{Perneczky2006}. These bands are shown in Table~\ref{tab:mmse_bands}. The MMSE is a 30-point screening test of global cognition (orientation, attention, memory, language, visuoconstructional ability). Lower MMSE scores reflect greater impairment~\cite{folstein1975mmse}. Note that inference is transcript-only: the test case’s MMSE is never provided, MMSE enters only through exemplar probabilities. 

\begin{table}[t]
  \centering
  \caption{Conventional MMSE bands used as anchors \cite{Perneczky2006}.}
  \label{tab:mmse_bands}
  \begin{tabular}{cc}
    \toprule
    \textbf{MMSE score} & \textbf{Category} \\
    \midrule
    30       & No impairment \\
    26--29   & Questionable \\
    21--25   & Mild \\
    11--20   & Moderate \\
    0--10    & Severe \\
    \bottomrule
  \end{tabular}
\end{table}

Let \(y\in\{0,1\}\) denote the example’s class (AD \(=1\), HC \(=0\)), and let \(m\) be the subject’s MMSE score (integer \(0\)–\(30\)). We elicit a proxy probability for AD that is monotonic in \(m\) and lies in \([0,1]\). AD examples map to \([0.50,1)\) (rising toward 1 as \(m\) decreases); HC examples always map below \(0.50\). Let \(\sigma(z)=1/(1+e^{-z})\). We map MMSE to an AD probability via
class-conditional sigmoids (Eqs.~(\ref{eq:ad_proxy2}) and~(\ref{eq:hc_proxy2})).
\begin{equation}
\label{eq:ad_proxy2}
P(\mathrm{AD}\mid y{=}1,m)=\sigma\!\Big(\frac{30-m}{T_{\mathrm{AD}}}\Big),
\end{equation}
\begin{equation}
\label{eq:hc_proxy2}
P(\mathrm{AD}\mid y{=}0,m)=\sigma\!\Big(\frac{30-m}{T_{\mathrm{HC}}}\Big)-0.5,
\end{equation}
with temperatures fixed by two anchor conditions at \(m{=}26\):
\[
P(\mathrm{AD}\mid y{=}1,26)=0.60,\qquad P(\mathrm{AD}\mid y{=}0,26)=0.40.
\]
These anchors center both curves at the “Questionable” boundary in Table~\ref{tab:mmse_bands} and determine \(T_{\mathrm{AD}}\) and \(T_{\mathrm{HC}}\) uniquely, without any fitting or post hoc calibration on ADReSS. The mapping yields visible separation near \(m\in[26,30]\), where most HC subjects lie. It intentionally assigns higher probabilities to HC as impairment increases (lower \(m\)), while keeping all HC scores below the decision threshold of \(0.50\). In practice, impaired HC cases receive probabilities near \(0.50\) rather than near \(0\), avoiding overly confident negatives for clinically ambiguous subjects. The goal from this mapping is tying the model’s score to a clinically familiar severity signal and enforcing a monotone relation. Using a numeric probability in $[0,1]$ rather than a high/low confidence tag preserves ranking information, enables threshold‐free discrimination metrics (AUC), and improves comparability.

In addition to the random nested interleave strategy, we evaluate term frequency–inverse document frequency (TF--IDF) interleave, which ranks training exemplars by cosine similarity between TF--IDF vectors and the test transcript, selects the top $k$ per class, then alternates AD and HC. Moreover, to test whether the effect is backbone-agnostic rather than Mistral-specific, we also apply the same MMSE-Proxy method to Qwen \cite{qwen3-8b-hf} and report the results.

\subsection{Reasoning-augmented examples with GPT-5}\label{sec:protocol2}
While experimenting with Mistral-7B-Instruct on Cookie Theft transcripts, we observed a structure-copying effect: when examples do not include rationales, the model often omits them at inference time even when explicitly requested, and may mirror placeholder text, especially if the number of examples increases beyond four. Because brief, transcript-grounded rationales improve interpretability for clinicians, we construct a new few-shot pool in which each example includes a concise comment plus an MMSE-anchored probability. We generate these examples with GPT-5 because it reliably follows the rationale constraint even with many heterogeneous examples. In addition, GPT-5 allows attaching the Cookie Theft image so the generator can combine visual and linguistic cues when drafting the example comment. This multimodal capability is used only during construction; evaluation remains strictly transcript-only.

For each training subject, GPT-5 receives: (i) the Cookie Theft image, (ii) the subject’s transcript, (iii) the MMSE score, and (iv) the true label, and must return valid JSON with exactly three fields: comment (brief reasoning, max 100 tokens), alzheimers\_prediction $\in \{\texttt{YES},\texttt{NO}\}$ as an exact match, and probability\_score $\in [0,1]$ that is consistent with the label and guided by the MMSE bands. We enforce probability $<0.5$ for \texttt{NO} and $>0.5$ for \texttt{YES}. MMSE influences the probability only, never the comment text. Since this example pool is generated by a different model (GPT-5) and uses image conditioning during construction, we report it separately. The pool was generated once and then frozen; no transcripts from the test split were used. At inference, the evaluated model receives transcripts only; neither the Cookie Theft image nor MMSE are provided. Examples are selected from this pool and appended to the prompt using the random nested interleave described in Section~\ref{sec:protocol0}.

\subsection{Models and evaluation}\label{sec:models_eval}

We use the open source model Mistral\cite{mistral7b} as our primary LLM because it was the best performing model on manual transcripts in previous work \cite{botelho2024macro}.
For MMSE-Proxy Prompting we use Mistral-7B-Instruct v0.2 to mirror \cite{botelho2024macro}. To check that the effect is not model-specific, we also evaluate Qwen3-8B under the same protocol \cite{qwen3-8b-hf}.
For Reasoning-augmented examples' prompting we use Mistral-7B-Instruct v0.3, selected for more reliable format adherence. 

 We parse the JSON and use the explicit 'alzheimers\_prediction' for accuracy. AUC is computed directly from 'probability\_score' with no training or post hoc calibration. We record evaluation results for 0 to 20 examples per class (up to 40 shots total), constrained by the context window. We use near-greedy sampling (temperature 0.01, top-$k$ 50, top-$p$ 1.0). Each experiment is repeated three times with different random seeds, and we report mean $\pm$ standard deviation (std) over the three runs.
 

\begin{table}[t]
\centering
\caption{Zero-shot and $k{=}14$ performance with and without the MMSE-Proxy (mean $\pm$ std across runs).}
\begin{tabular}{lcc}
\toprule
Setting & Accuracy & AUC \\
\midrule
Zero-shot ($k{=}0$)            & $0.65\,\pm\,0.00$ & $0.63\,\pm\,0.00$ \\[0.25em]
No-Proxy baseline ($k{=}14$)   & $0.76\,\pm\,0.01$ & \multicolumn{1}{c}{---} \\[0.25em]
MMSE-Proxy ($k{=}14$)    & $0.82\,\pm\,0.01$ & $0.86\,\pm\,0.01$ \\
\bottomrule
\end{tabular}
\label{tab:results_k}
\end{table}

\begin{table}[t]
\centering
\caption{Reasoning-augmented examples: zero-shot and $k{=}19$ performance
(mean $\pm$ std across runs).}
\begin{tabular}{lcc}
\toprule
Setting & Accuracy & AUC \\
\midrule
Zero-shot ($k{=}0$) & $0.54\,\pm\,0.00$ & $0.66\,\pm\,0.01$ \\[0.4em]
$k{=}19$ & $0.82\,\pm\,0.01$ & $0.83\,\pm\,0.01$ \\
\bottomrule
\end{tabular}
\label{tab:reasoning}
\end{table}


\begin{figure}[t]
\centering
\includegraphics[width=\columnwidth]{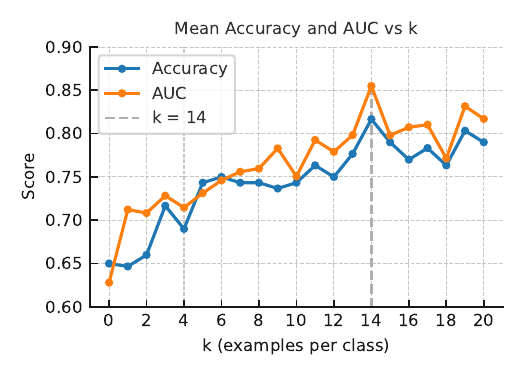}
\caption{MMSE-Proxy prompting evaluation: Mean accuracy and AUC as a function of $k$.}
\label{fig:results_k}
\end{figure}

\begin{figure}[t]
\centering
\includegraphics[width=\columnwidth]{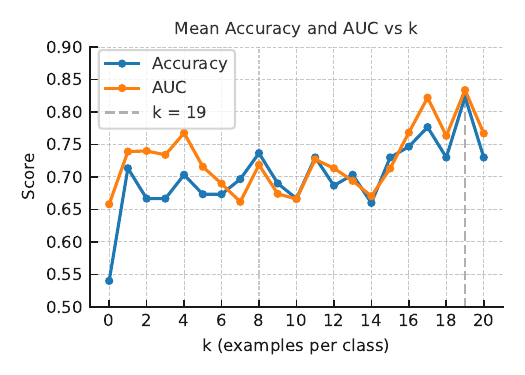}
\caption{Reasoning augmented examples prompting: Mean accuracy and AUC as a function of $k$.}
\label{fig:acc_reason}
\end{figure}


\section{Results and Discussion}\label{sec:results}
All runs yielded zero parsing failures. Subsection \ref{res1} presents results with MMSE-Proxy prompting, then subsection \ref{res2} reports results of prompting using reasoning-augmented examples with GPT-5. Finally, we discuss the results and compare them with previous work (subsection \ref{discussion}).

\subsection{MMSE-Proxy Prompting}\label{res1}
In the MMSE-Proxy Prompting experiment with random nested interleave example selection, performance increases as $k$ grows and plateaus in the mid-range (Figure~\ref{fig:results_k}). The highest scores are achieved at $k{=}14$ with 0.82 accuracy and 0.86 AUC and narrow variance (Table~\ref{tab:results_k}). Under the same instruction with $k{=}0$, the zero-shot baseline is 0.65 accuracy and 0.63 AUC. These gains show that exposing the model to class-balanced demonstrations improves discrimination without any parameter updates.

To isolate the effect of the MMSE-proxy, we ablate it and re-run the protocol with the same seeds using random nested interleave. At \(k{=}14\), the no-proxy baseline achieved \(0.76 \pm 0.01\) accuracy, compared with \(0.82 \pm 0.01\) for the MMSE-proxy under identical settings (Table \ref{tab:results_k}). This paired comparison at fixed \(k\) indicates that the improvement is due to probability anchoring rather than seed variance or exemplar count. Note that AUC is not reported for the no-proxy baseline because removing probability-formatted demonstrations led the model to omit the probability field at inference, despite explicit instructions, yielding incomplete outputs that prevent consistent AUC computation; however, accuracy is unaffected.
We additionally evaluate a retrieval-based variant where examples are selected by TF-IDF cosine similarity to the test transcript instead of using a single global examples set for all test transcripts as in random nested interleave.  TF-IDF interleave follows a similar trajectory where scores increase as $k$ increases and peak at 0.79 accuracy and 0.83 AUC at $k{=}17$. This indicates that mid-range balanced in-context supervision drives the strongest gains, while TF-IDF retrieval does not surpass random nested interleave in performance. Applying the same protocol with random nested interleave to Qwen yields the same trend, peaking at 0.76 accuracy and 0.82 AUC at $k{=}11$.

\subsection{Reasoning-augmented examples with GPT-5}\label{res2}
Using demonstrations constructed by GPT-5, the evaluated Mistral model reliably emits concise comments together with labels and probabilities. Accuracy and AUC rise with $k$ and peak at $k{=}19$ with 0.82 accuracy and 0.83 AUC (Figure~\ref{fig:acc_reason}). Note that the zero-shot results in Table~\ref{tab:reasoning} correspond to the Mistral v0.3, which explains the difference from the MMSE-Proxy zero shot result. By introducing 19 examples per class, the model accuracy increases from 0.54 to 0.82, which is the highest accuracy reported for error-free few-shot prompting using Mistral on the ADReSS dataset so far. These results indicate that including short, transcript-grounded rationales in the demonstrations can improve format adherence and interpretability while maintaining competitive performance.

\subsection{Discussion}\label{discussion}
 
Simple, class-balanced few-shot prompting improves Alzheimer’s detection from Cookie Theft transcripts without any parameter updates. Few-shot prompting provides clear gains over zero-shot baselines and enables reliable output without training or calibration. The forced-decision instruction avoids artifacts introduced by failure conventions \cite{botelho2024macro} and allows fair comparison.
MMSE-proxy prompting leads to an increased AUC score reaching 0.86. With random nested interleave, accuracy and AUC rise with $k$ and stabilize in the mid-range. The ablation at $k{=}14$ shows a +6 point accuracy gain over the no-proxy baseline, suggesting that anchoring demonstrations to a clinically meaningful severity signal helps the model separate borderline cases while keeping output format stable.
The effect replicates on Qwen, which suggests the protocol is not tied to a specific backbone model. The comparison between selection strategies shows that retrieving semantically similar examples to the test transcript does not necessarily improve the results, implying that balanced supervision and preserved ordering matter more than nearest-neighbor retrieval. Nonetheless, TF-IDF interleave offers a deterministic, reproducible alternative with good performance.
Reasoning-augmented examples help the model adhere fully to the required JSON format while maintaining competitive performance. They also improve interpretability by providing a brief reason for each decision taken by the LLM, which can aid clinicians in their diagnosis. Overall, these prompting methods are computationally light and competitive with recent prompting baselines on the ADReSS dataset using Mistral (Table~\ref{tab:prompting_compact_numbers}).

\begingroup
\renewcommand{\arraystretch}{1.15}
\small
\begin{table}[t]
\centering
\caption{Accuracy and AUC (mean $\pm$ std over 3 runs) for prompting-only
methods on ADReSS test (Mistral). $k$ is the number of few-shot examples per class.}
\begin{tabular}{lccc}
\toprule
Prompting Method & $k$ & Accuracy & ROC--AUC \\
\midrule
MMSE-Proxy (ours)$^{*}$          & 14 & 0.82 $\pm$ 0.01 & 0.86 $\pm$ 0.01 \\
Reasoning-augmented (ours)$^{**}$& 19 & 0.82 $\pm$ 0.01 & 0.83 $\pm$ 0.01 \\
Delta-KNN~\cite{li2025delta}$^{**}$ & 2  & 0.76 $\pm$ 0.01 & 0.85 $\pm$ 0.03 \\
Botelho et al.~\cite{botelho2024macro}$^{*}$ & 0 & 0.75 & -- \\
\bottomrule
\multicolumn{4}{l}{\footnotesize $^{*}$ Mistral-7B-Instruct v0.2 \quad
$^{**}$ Mistral-7B-Instruct v0.3}
\end{tabular}
 \label{tab:prompting_compact_numbers}
\end{table}
\endgroup

\section{Conclusion}

We present a simple, controlled few-shot prompting protocol with nested interleave for AD detection from Cookie Theft transcripts on the ADReSS dataset that is accurate and reproducible without parameter updates. To our knowledge, this is the first study to anchor elicited probabilities to MMSE bands so that AUC can be reported , and to leverage a multimodal construction step for examples while keeping evaluation strictly text-based. 0.82 accuracy and 0.86 AUC with zero parsing failures are among the strongest prompting results on ADReSS under transcript-only evaluation. Future work will incorporate audio or other modalities, explore more principled exemplar selection, and extend beyond English using cross-lingual Cookie Theft resources.

\section{COMPLIANCE WITH ETHICAL STANDARDS}
This work uses the ADReSS 2020 subset of DementiaBank (TalkBank) under its data-use terms.

\bibliographystyle{IEEEbib} 








\end{document}